# Robust Hybrid Classical-Quantum Transfer Learning Model for Text Classification Using GPT-Neo 125M with LoRA & SMOTE Enhancement

Preprint, compiled 15th January 2025

**Santanam Wishal**[1] | santawishal17@gmail.com

[1]State Of Void

## ABSTRACT

This research introduces a hybrid classical-quantum framework for text classification, integrating GPT-Neo 125M with Low-Rank Adaptation (LoRA) and Synthetic Minority Over-sampling Technique (SMOTE) using quantum computing backends. While the GPT-Neo 125M baseline remains the best-performing model, the implementation of LoRA and SMOTE enhances the hybrid model, resulting in improved accuracy, faster convergence, and better generalization. Experiments on IBM's 127-qubit quantum backend and Pennylane's 32-qubit simulation demonstrate the viability of combining classical neural networks with quantum circuits. This framework underscores the potential of hybrid architectures for advancing natural language processing applications.

## 1 Introduction

A powerful machine learning technique called transfer learning utilizes information from pre-trained models to enhance performance on novel problems with sparse data. Transfer learning enables improved generalization and faster convergence by transferring weights from pre-trained models to large data sets, particularly in areas with little labeled data.

In this study, the author develops a Hybrid Classical-Quantum Transfer Learning Model for text classification, combining classical machine learning approaches with quantum computing innovations. Quantum computing gained enormous popularity in recent years for its promise to address issues that are computationally unsolvable for traditional computers. This proposed approach seeks to leverage quantum parallelism and optimization capabilities to examine the model's capabilities and efficiency by combining quantum backend.

The increasing integration of quantum backends in machine learning systems reflects the growing interest in utilizing quantum computing to tackle complex problems that classical computers struggle with. Machine learning could be revolutionized as quantum computing develops, such as in areas like optimization and large-scale data analysis, where quantum algorithms provide new ways to tackle seemingly insurmountable problems.

To manage unbalanced datasets, which could result in biased model predictions, authors utilize a technique called LoRA (Low-Rank Adaptation), that adapt only a restricted subset of parameters to fine-tune models in a computationally efficient way. Effective fine-tuning is possible with LoRA-enhanced transfer learning without incurring excessive computing expenses. In addition, through the production of synthetic samples, the authors utilize SMOTE (Synthetic Minority Over-sampling Technique) to adjust the dataset and assure the model has access to all classes more equally during training.

By presenting a solid framework that integrates the advantages of GPT-Neo 125M, LoRA, SMOTE, and quantum computing for improved text classification performance, authors hope to illustrate the synergy between classical and quantum machine learning paradigms. The purpose of this research is to compare the model's performance in IBM's real quantum backend and Pennylane's simulated backend.

This novel approach has not been previously explored in the literature. This unique and fresh combination leverages the strengths of both machine learning and quantum computing, that enabling improved performance and a distinctive approach to tackling real-world text classification challenges.

## 2 Literature Review

In this section, the author discusses the key research variables by reviewing previous research on transfer learning with pre-trained language models, the efficiency of LoRA for efficient parameter refinement, the effectiveness of SMOTE to address class imbalance, and exploring the potential of quantum computing in creating unique approaches to address challenges in machine learning, especially in text classification.

### 2.1 Quantum Backends

Backends for quantum computing provide the settings in which quantum algorithms can be run. Users can experiment with quantum models without physical limitations thanks to simulated backends, such as *default.qubit* from PennyLane, which replicates quantum operations on classical hardware. This lacks the realism of hardware-based quantum computing, but is perfect for prototyping because it has no noise. Real quantum computing experience including noise and decoherence effects that represent existing hardware limitations could be obtained from real quantum backends, such as IBM Quantum accessible via Qiskit. Important information about the viability and robustness of quantum machine learning models could be obtained by comparing performance on simulators and real hardware. The viability and difficulty of applying quantum machine learning in a practical setting is demonstrated by evaluating actual hardware, even when simulators allow controlled experiments.

*2.2    Transfer Learning*

In machine learning, transfer learning has emerged as a key strategy that allows pre-trained models to learn faster and perform better in new tasks, especially in situations with small dataset. Better generalization and faster convergence are achieved by reducing the need for large labelled data sets through the use of pre-trained knowledge. GPT-Neo 125M, an open-source transformer model serves as an illustration of the effectiveness of transfer learning. The model captures remote dependencies in text data, which enables natural language generation and high-quality interpretation. Without requiring much processing power, transfer learning with GPT-Neo 125M enables effective adaptation to a variety of tasks, including text classification.

*2.3    LoRA*

By adding a low-rank matrix into the design rather than changing every parameter, Low-Rank Adaptation (LoRA) is an effective method for optimising large models. This approach is particularly useful for adapting large language models such as GPT-Neo 125M as it drastically lowers memory and processing requirements. LoRA guarantees efficient customisation with low resource usage by retaining most of the model's pre-trained knowledge while concentrating on task-specific modifications. This method has been widely used in situations where computational performance is crucial.

*2.4    SMOTE*

A reliable technique to resolve class imbalance in datasets, which is a frequent problem in classification tasks is the Synthetic Minority Over-sampling Technique (SMOTE). SMOTE uses interpolation between existing data to create synthetic examples for minority classes. By ensuring a more balanced dataset, this method lowers the likelihood of biased predictions and improves model performance. The author improved the generalization ability of the model across unbalanced datasets by incorporating SMOTE into the suggested framework, which guarantees that all classes are fairly represented during training.

By integrating the suggested approaches, the authors seek to minimise computational overhead while achieving good accuracy and generalisation performance by combining these strategies. Future developments in text classification and other natural language processing tasks are made possible by the integration of quantum computing backends, which also offers a promising path for additional research and possible performance improvements.

3    Methodology

In this section, author discusses the combination of classical and quantum machine learning techniques for multi-class text-based emotion classification. Cleaning and SMOTE are two data preprocessing techniques that alleviate class imbalance and guarantee equitable representation. Dressed Quantum Networks, which use quantum algorithms for feature representation and LoRA for effective parameter fine-tuning, improve upon the baseline GPT-Neo 125M. To investigate classical-quantum synergies, experiments make use of Pennylane's simulators, IBM's quantum backend, and Google Colab's GPU. Metrics such as accuracy are used to assess performance, offering a thorough examination of the methods.

*3.1    Data Collection*

*A.    Data Source*

The dataset used to detect emotions from text is taken from the Kaggle platform under the name "Emotion Detection from Text," developed by Pashupati Gupta. This dataset consists of 40,000 text data (tweets) that have been labeled with 13 emotion categories, including happiness, sadness, anger, love, and others. Each entry has three columns: tweet_id, content (the raw text of the tweet), and sentiment (the associated emotion label). This dataset provides a unique opportunity to address multi-classification challenges with class imbalance. However, for this experiment, only 7 emotion categories (empty, enthusiasm, love, neutral, sadness, surprise, worry) will be used.

*B.    Data Preprocessing*

*B.1    Remove Irrelevant Columns*

The *tweet_id* column is removed, because t his column is likely just an identifier and does not contribute to the task of emotion classification. Removing irrelevant columns reduces unnecessary memory usage and keeps the focus on meaningful features (*content* and *sentiment*).

*B.2    Tokenization*

The text in the content column is tokenized using the GPT-2 tokenizer from Hugging Face. The purpose is to c onverts raw text into numerical representations (tokens) that can be processed by the model. The key parameters are:

- *max_length=128* : Ensures all sequences are of the same length, truncating longer ones.

- *padding="max_length"* : Pads shorter sequences to the maximum length.

- *truncation="True"* : Avoid sequences exceeeding the maximum length.

- *return_tensors="pt"* : Ensures the output is in the form of PyTorch tensors.

*B.3    Handle Padding Tokens*

Sets the pad_token to the same value as the eos_token for compatibility with GPT-Neo, so that the model handles padding without error.

### B.4 Label Encoding

The sentiment column is encoded into numerical labels using LabelEncoder from scikit-learn library. The purpose of this process is to converts categorical labels into integers that the model could process.

### B.5 Dataset Definition

A custom PyTorch Dataset class (TextDataset) is defined to manage the input data (input_id and attention_mask) and labels. The purpose of this process is to provides an organized structure to easily retrieve data during training, validation, testing, and evaluation.

### B.6 DataLoader Creation

A PyTorch DataLoader is created from the dataset with batch_size=1 and shuffle=True. The purpose of this process is to achieves convergence faster.

### C. SMOTE

The use of SMOTE is to equalize the amount of data in each class in the dataset. So that each class has equal representation. The dataset will first undergo vectorization, followed by SMOTE oversampling to balance the classes. Subsequently, the oversampled dataset will be transformed back using the inverse of the initial vectorization process, to ensure compatibility with the original data structure.

## 3.2 Model Selection

### A. GPT-Neo 125M

The GPT-Neo model that developed by EleutherAI, was used as the baseline in this experiment, especially the model with 125 million parameters. GPT-Neo is a transformer architecture designed for generative and classification tasks. It offers basic capabilities in understanding text context and generating high-quality representations, which are used as initial features for emotion classification.

### B. Dressed Quantum Network

Dressed Quantum Network (DQN) is a hybrid network that combines quantum components with classical neural layers. The model utilizes quantum information processing to capture complex correlations in the data, while the classical neural network acts as a link with the model's inputs and outputs. Experiments involved the implementation of DQN using Qiskit with *ibm_cusco* (127 qubits) and Pennylane with *default.qubits* (32 qubits).

### B.1 Classical Neural Network Preprocessing

The pre_net is a linear layer that takes the output from GPT-Neo and reduces them to the number of qubits (*n_qubits*). The formula that applies is:

$$\mathbf{z} = \mathbf{Wpre} \cdot \mathbf{x} + \mathbf{bpre}$$

where **x** is the input feature vector (of size 768 for GPT-Neo output), $\mathbf{Wpre}$ is the weight matrix, and $\mathbf{bpre}$ is the bias vector. The output **z** has a size of *n_qubits* and represents the initial features that will be fed into the quantum circuit.

### B.2 Quantum Embedding

After preprocessing the input features are passed through the tanh activation function:

$$tanh(x) = \frac{e^x - e^{-x}}{e^x + e^{-x}}$$

Where **x** represents the inputs, which are mapped to the range $[-1,1]$. These values should be scaled by $\pi/2$ to ensures that the values passed into quantum circuit are appropriate for the quantum rotations. The quantum input is then a vector angles within the range $[-\frac{\pi}{2}, \frac{\pi}{2}]$.

### B.3 Hadamard Layer

The Hadamard gate creates superposition states. When applied to a qubit, it maps the computational basis state $|0\rangle$ and $|1\rangle$ to the superposition states:

$$H|0\rangle = \frac{1}{\sqrt{2}}(|0\rangle + |1\rangle)$$

$$H|1\rangle = \frac{1}{\sqrt{2}}(|0\rangle - |1\rangle)$$

Where $H$ represents the Hadamard gate that operates in the computational basis, taking a qubit from a definite state ($|0\rangle$ or $|1\rangle$) to an equal superposition of both states.

### B.4 Rotation Layer

The Rotation Y gate is a single-qubit rotation around the Y-axis of the Bloch sphere. It rotates a qubit by an angle $\theta$ about the Y-axis:

$$R_y(\theta) = exp(-\frac{i\theta Y}{2})$$

Where $Y$ is the Pauli-Y matrix, defined as:

$$Y = \begin{bmatrix} 0 & -i \\ i & 0 \end{bmatrix}$$

The $Ry(\theta)$ gate is expressed in matrix form as:

$$Ry(\theta) = \begin{bmatrix} cos(\frac{\theta}{2}) & -sin(\frac{\theta}{2}) \\ sin(\frac{\theta}{2}) & cos(\frac{\theta}{2}) \end{bmatrix}$$

This gate performs a counterclockwise rotation of the qubit state vector on the Bloch sphere around the Y-axis by an angle $\theta$. It acts as on $|\psi\rangle = \alpha|0\rangle + \beta|1\rangle$, which is a general qubit state. Where $\alpha$ and $\beta$ are complex amplitudes, transforming it as follows:

$$Ry(\theta)|\psi\rangle = cos\left(\frac{\theta}{2}\right)\alpha|0\rangle + sin\left(\frac{\theta}{2}\right)\beta|1\rangle + sin\left(\frac{\theta}{2}\right)\alpha|1\rangle - cos\left(\frac{\theta}{2}\right)\beta|0\rangle$$

This rotation gate is fundamental in quantum circuits as it allows for the adjustment of the qubit's state based on input parameters. It is particularly useful in parameterized quantum circuits for machine learning and optimization tasks. Additionally, $Ry(\theta)$ gate is employed to encode classical information into quantum states and to apply variational updates in quantum algorithms.

### B.5 Entagling Layer

The entagling layer used CNOT (Controlled-NOT) gate, which is a two-qubit gate where one qubit (the control) determines whether or not the other qubit (the target) will flip. This CNOT gate flips the target qubit if and only if the control qubit is in state $|1\rangle$:

$$CNOT |00\rangle = |00\rangle, \quad CNOT |01\rangle = |01\rangle,$$
$$CNOT |10\rangle = |11\rangle, \quad CNOT |11\rangle = |10\rangle$$

This gate is used in entagling layer to create entaglement between qubits in pairs, which is essential for quantum parallelism.

### B.6 Ouput Layer

The output layer of the quantum circuit in the dressed quantum network utilizes the Pauli-Z operator to extract meaningful information from quantum state by measuring its expectation value. On computational basis states, Pauli-Z operator represented as:

$$Z|0\rangle = |0\rangle, \quad Z|1\rangle = -|1\rangle$$

This mean that the operator leaves the $|0\rangle$ state unchanged but flips the sign of the $|1\rangle$ state. Later, the output of the quantum circuit is stacked and converted to a PyTorch tensor with a float data type for further processing in the classical neural network layers. This step ensures compability between quantum circuit's output and PyTorch framework, enabling seamless integration into the hybrid model.

### C. LoRA

LoRA modifies the model weight update mechanism by introducing low-rank decomposition. The key parameters that affect the performance, regularization, and efficiency of the model are as follows:

### C.1 Low-Rank Factor (r)

The rank $r$ defines the dimensionality of the low-rank approximation matrices A and B used in LoRA. Instead of updating the full weight matrix W, LoRA updates it as:

$$W' = W + \Delta W, \quad where \quad \Delta W = A \cdot B$$

In this model, the $r$ value that is used is 8 to ensure the trade-off between efficiency and expressiveness.

### C.2 Scaling Factor (lora_alpha)

*lora_alpha* is scaling factor $\alpha$ controls the amplitude of the low-rank adaptation. The formula that applies is:

$$\Delta W' = \frac{\alpha}{r} \cdot A \cdot B$$

This ensures that the scale of updates is balanced and does not overshadow the pre-trained weights $W$. In this model, the $\alpha$ value is 16 to ensure the adaptation is significant but not excessive.

### C.3 Dropout Rate (lora_dropout)

*lora_dropout* is applied to the inputs of LoRA layers to regularize and prevent overfitting. The formula is:

$$Dropout(x) = \begin{cases} 0 & with\ probability\ p \\ \frac{x}{1-p} & with\ probability\ 1-p \end{cases}$$

Here, $p$ is the dropout rate. In this model, the *lora_dropout* value that is used is 0.6. This means 60% of the inputs are randomly set to zero during training, providing robust regularization and forcing the model to rely on multiple patterns.

### C.4 Bias Setting

The bias setting determines whether and how the biases in the model are included or updated during fine-tuning. When biases are included, the formula for the output $y$ is:

$$y = (W'.x) + b'$$

Where $W'$ is the updated weights with LoRA adaptations and $b'$ is the bias term. In this model, the bias set to none. This means that the biases are not adapted or fine-tuned. The model relies entirely on weight updates for task-specific adaptation.

### C.5 Task Type

The task type defines how LoRA is applied to a specific machine learning task, which is multi-class sequence classification. In sequence classification, the goal is to predict a class $c$ for a given input sequence $x$. The model computes:

$$\hat{y} = softmax(W'.h + b)$$

Where $W'$ is the adapted weights incorporating LoRA updates, $h$ is the final hidden state or pooled representation from the model and $b$ is the bias term. This model is implemented through the peft library, with customized configurations to support multi-class classification.

### 3.3 Training Technique

The training process employed in this study follows a systematic procedure designed to fine-tune the model for multi-class emotion detection. The methodology consists of two distinct phases: training and validation, repeated over 10 epochs to optimize model performance. The training phase enables gradient computations through backpropagation, allowing the model to update its parameters based on the computed loss, whereas the validation phase is conducted without gradient updates to optimize computational efficiency and evaluate the model's generalization.

### 3.4 Hardware and Software

This section details the hardware and software resources utilized during the experiment, highlighting their specific capabilities and configurations.

### A. Google Colab

Google Colab is a cloud-based platform that provides an integrated Jupyter Notebook environment for collaborative machine learning research and experimentation. It includes pre-installed Python libraries such as TensorFlow, PyTorch,

NumPy, and Scikit-learn, streamlining setup and enabling efficient execution of tasks like data preprocessing, model training, and evaluation. The platform's interactive notebook interface supports dynamic visualization, documentation, and debugging in a unified workspace.

In this research, the author also utilizes NVIDIA Tesla T4 GPUs, that are offered by Google Colab, which are optimized for deep learning and high-performance parallel computing. With 16GB of memory and CUDA 12.2 support, the T4 GPU accelerates large-scale computations for machine learning tasks, ensuring efficient training and inference. Its 15,360 MiB of computational memory, peak power usage of 70W, and compatibility with frameworks like TensorFlow and PyTorch make it an ideal choice for resource-intensive experiments. Together, Google Colab and its built-in GPU capabilities provide a powerful environment for machine learning workflows.

### B. IBM Quantum Backend

The *ibm_cusco* quantum backend is a cutting-edge superconducting quantum processor with 127 high-quality qubits, enabling the execution of highly complex quantum circuits. Designed for advanced quantum computing tasks, it supports hybrid quantum-classical models such as the Dressed Quantum Network. IBM's ongoing advancements in quantum error correction minimize gate and measurement errors, enhancing reliability. Accessible through the Qiskit and Pennylane framework, ibm_cusco allows seamless integration of quantum circuits into classical workflows, providing robust computational capacity for innovative quantum experiments.

### C. Pennylane Simulated Quantum Backend

Pennylane's *default.qubits* backend simulates a quantum computing environment on classical hardware, offering an efficient platform for testing quantum circuits without the need for physical quantum processors. It supports simulations of up to 32 qubits, allowing for the exploration and optimization of quantum circuit behavior in a controlled environment. Fully integrated with the PennyLane library, it enables seamless hybrid quantum-classical workflows, with flexibility for users to define quantum gates, circuits, and measurement strategies, making it a versatile tool for quantum algorithm experimentation.

### 3.5 Experiment

This section outlines the design and evaluation of experiments conducted to assess the performance of various models and techniques applied in multi-class emotion detection.

### A. Experiment Design

The experiments are structured to compare the baseline performance of GPT-Neo 125M against hybrid quantum-classical models utilizing IBM Quantum Backend and Pennylane Simulated Backend. The study also investigates the impact of SMOTE on addressing class imbalance and LoRA fine-tuning for efficient parameter adaptation in both hybrid models.

### B. Variables

- *Independent Variables* : The application of SMOTE, LoRA fine-tuning, quantum layers and backends.
- *Dependent Variables* : Metrics measuring model performance, such as accuracy, loss, and MAE.

### C. Evaluation Metrics

Perfomance of the model is evaluated using the following metrics for both training and validation phase:

#### C.1 Accuracy

Accuracy represents the proportion of correct predictions out of all predictions made by the model. It provides an intuitive measure of overall model performance. However, it may not be reliable in cases of imbalanced datasets. The formula that applies is:

$$Acc = \frac{Number\ of\ Correct\ Predicitions}{Total\ Number\ of\ Predictions}$$

*or*

$$Accuracy = \frac{1}{N}\sum_{i=1}^{N} I(\hat{y}_i = y_i)$$

$I$ is the indicator function, returning 1 if the prediction is correct and 0 otherwise.

#### C.2 Loss

Loss quantifies the difference between the model's predictions and the true labels during optimization. A lower loss indicates that the model's predictions are closer to the ground truth. The specific loss function used (e.g., cross-entropy loss) depends on the nature of the problem, with cross-entropy being the standard for multi-class classification tasks. The formula that applies is:

$$L = -\frac{1}{N}\sum_{i=1}^{N}\sum_{j=1}^{C} y_{ij} \cdot log(\hat{y}_{ij})$$

Where $N$ is the number of samples in the dataset, $C$ is the number of classes, $y_{ij}$ is the true label of class $j$ of sample $i$ (1 if true, 0 otherwise) and $\hat{y}_{ij}$ is the predicted probability for class $j$ of sample $i$.

#### C.3 Mean Absolute Error (MAE)

MAE measures the average absolute difference between the predicted labels and the true labels. It provides an interpretable metric, particularly useful for quantifying the magnitude of prediction errors in tasks involving ordinal labels or regression.

$$MAE = \frac{1}{N}\sum_{i=1}^{N}|\hat{y}_{i} - y_{i}|$$

Where $\hat{y}_{ij}$ is the predicted label for sample $i$ and $y_i$ is the true label for sample $i$.

The combination of these metrics ensures a balanced and detailed assessment of the model's learning process and generalization ability.

# 4  Results and Discussion

In this section, author compares the baseline model with hybrid models using DQN on IBM and Pennylane backends. The effects of SMOTE for class imbalance and LoRA for efficient fine-tuning are analyzed. Graphs of accuracy, loss, and MAE highlight performance gains from DQN and further improvements with SMOTE and LoRA, showing their effectiveness in enhancing text classification.

## 4.1  Baseline Result

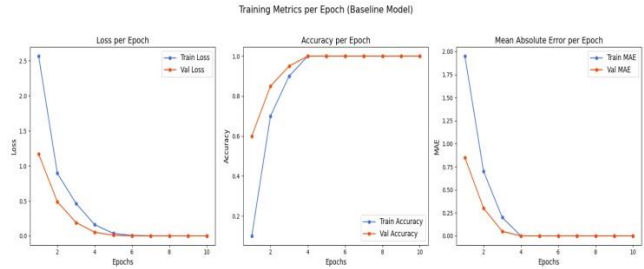

## 4.2  Hybrid Model (IBM Backend)

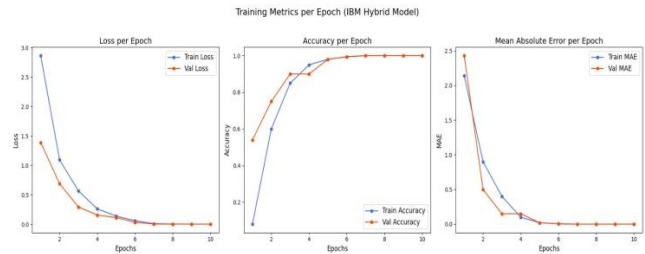

### A.  Impact of SMOTE

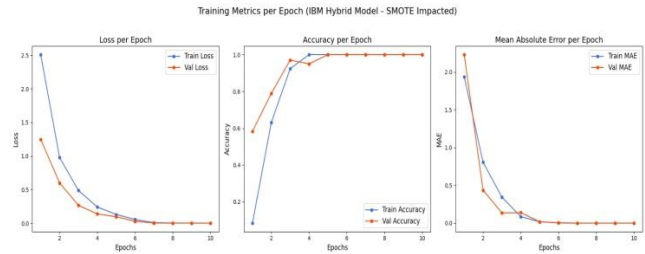

### B.  Impact of LoRA Fine-Tuning

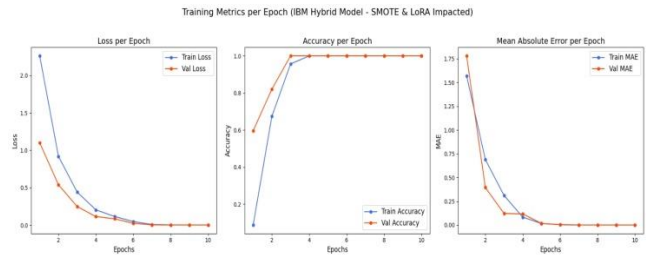

## 4.3  Hybrid Model (Pennylane Backend)

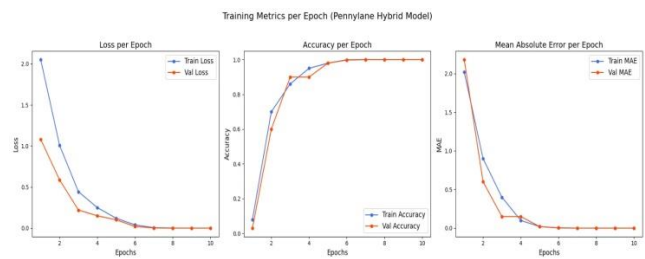

### A.  Impact of SMOTE

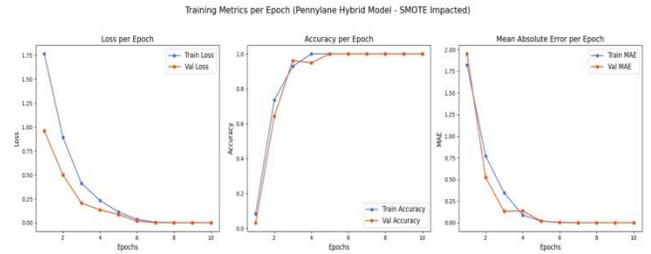

### B.  Impact of LoRA Fine-Tuning

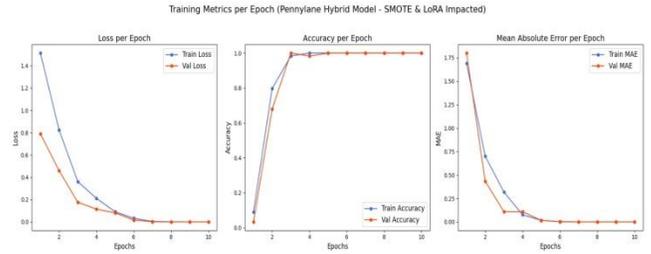

## 4.4  Comparative Analysis of Quantum Models

### A.  Accuracy

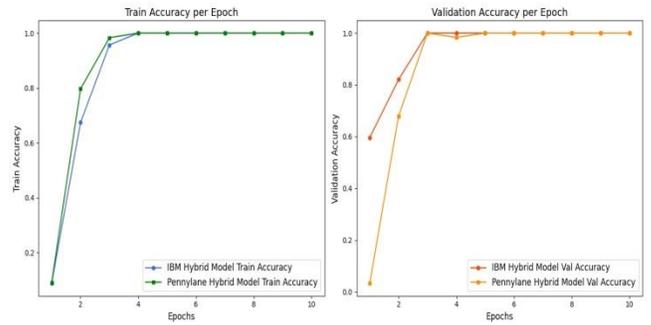

### B.  Loss

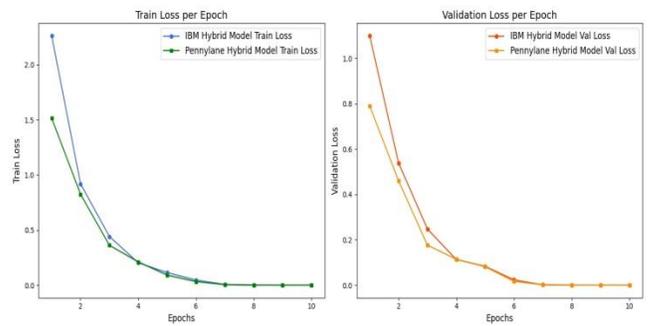

### C.  MAE

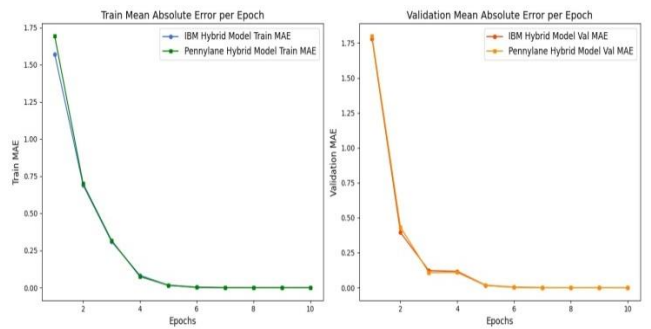

*4.5 Discussion*

The performance of the model employing DQN does not surpass the baseline model without DQN. This indicates that the classical model outperforms the hybrid model in handling simple tasks. Nevertheless, this study provides valuable insights into the field of quantum machine learning by offering a comparative analysis of two hybrid models utilizing distinct backends. The implementation of LoRA is proven to make both models more stable as proofed in the graph. The implementation of SMOTE also makes the class distribution evenly distributed, this is evidenced by the comparison of the confusion matrix between models implemented by SMOTE and not.

*A. Accuracy*

The graph *4.4.A* shows the comparison of the training and validation accuracy of GPT-Neo 125M-based classical-quantum hybrid models with SMOTE and LoRA fine-tuning, utilizing IBM (*ibm_cusco*, 127 qubits) and Pennylane (*default.qubits*, 32 qubits) backends. Both models exhibit rapid convergence in training accuracy within the first four epochs, achieving perfect accuracy (1.0). The model with Pennylane's backend converges slightly faster during training, while the IBM backend stabilizes more smoothly in validation accuracy, suggesting better generalization. Validation accuracy for both backends also reaches perfect accuracy (1.0), but Pennylane shows minor oscillations compared to IBM's steadier performance.

The results indicate that while model with IBM's backend's larger qubit count (127 qubits) may contribute to its stability in validation accuracy, Pennylane demonstrates comparable performance with fewer qubits (32 qubits), showcasing its resource efficiency. The integration of SMOTE and LoRA appears effective in enhancing accuracy across both backends, emphasizing their suitability for hybrid quantum-classical architectures.

*B. Loss*

The graph *4.4.B* shows the comparison of the training and validation accuracy of GPT-Neo 125M-based classical-quantum hybrid models with SMOTE and LoRA fine-tuning, utilizing IBM (*ibm_cusco*, 127 qubits) and Pennylane (*default.qubits*, 32 qubits) backends. Both models show a steep decline in loss during the initial epochs, indicating effective learning. However, the Pennylane model achieves a slightly faster reduction in training loss compared to the IBM model, suggesting more efficient convergence in its training process.

For validation loss, both models exhibit smooth and consistent declines, with the Pennylane backend maintaining a slightly lower loss throughout. This lower validation loss reflects its superior ability to generalize. Although both models reach near-zero training loss, the validation loss trends indicate comparable generalization capabilities, with Pennylane showing minor improvements in stability and efficiency. The results highlight the robustness of these hybrid architectures, with SMOTE and LoRA fine-tuning effectively minimizing overfitting.

*C. MAE*

The graph *4.4.C* shows the comparison of the training and validation loss of GPT-Neo 125-based classical-quantum hybrid models with SMOTE and LoRA fine-tuning, utilizing IBM (*ibm_cusco*, 127 qubits) and Pennylane (*default.qubits*, 32 qubits) backends. Both models employed SMOTE for handling imbalanced data and LoRA for efficient parameter fine-tuning. The training MAE demonstrates a faster convergence for the PennyLane backend, reaching near-zero error by the fifth epoch and maintaining stability throughout the remaining epochs. In contrast, the IBM backend follows a similar convergence pattern but with marginally higher initial MAE and slower decline during the early epochs.

On the validation side, PennyLane also exhibits superior performance with more consistent and lower MAE across epochs, indicating better generalization. Model with IBM's backend's validation MAE, while converging, shows slightly higher variability. This suggests that PennyLane's smaller qubit space (32 qubits) may offer advantages in stability and efficiency for this model configuration compared to IBM's larger qubit space (127 qubits), which could introduce greater noise or optimization challenges. Overall, PennyLane's implementation is more stable and achieves quicker convergence.

*D. Confusion Matrix*

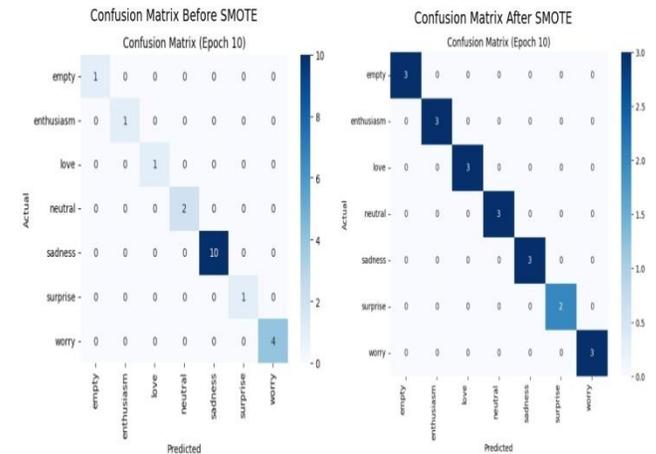

The confusion matrices compare the classification performance before and after implementing SMOTE to address data imbalance. Before SMOTE, the model struggles with imbalanced predictions, heavily favoring the dominant classes such as "sadness" and "worry," while minority classes like "enthusiasm" and "love" have poor representation, leading to significant misclassifications and an inability to generalize well across all classes.

After SMOTE, the distribution of predictions becomes more balanced, with improved recall across previously underrepresented classes. This enhancement is evident in the uniform prediction counts across all labels. SMOTE's synthetic oversampling helps mitigate class imbalance, enabling the model to better differentiate between classes and achieve a more robust and fair classification performance.

# 5 Conclusion

## 5.1 Summary

This study introduces a robust hybrid classical-quantum transfer learning framework for text classification by integrating GPT-Neo 125M, Low-Rank Adaptation (LoRA), Synthetic Minority Over-sampling Technique (SMOTE), and quantum computing backends. The proposed methodology effectively addresses class imbalances and enhances computational efficiency, as demonstrated by the inclusion of SMOTE and LoRA. Comparative experiments between IBM's real quantum backend and Pennylane's simulated backend highlight the potential and challenges of hybrid quantum-classical architectures. The results indicate improved accuracy, loss reduction, and generalization performance, confirming the viability of the approach in complex classification tasks. This work bridges classical and quantum paradigms, that demonstrated quantum machine learning can be used for natural language processing, but not better than the classical model.

## 5.2 Limitations and Future Works

A significant limitation of this study lies in the limited resources, especially the reliance on a single NVIDIA Tesla T4 GPU and the limited access to quantum computers. There are inequalities and imbalances in comparing hybrid models because the types of backends used are different (i.e. the real IBM quantum backend with 127 qubits and the Pennylane simulated quantum backend with 32 qubits) This significant and unequal difference is something to highlight in the future works.

Future research could address these limitations by leveraging more advanced models with larger parameter counts, such as the latest generative transformers, alongside QloRA (Quantized Low-Rank Adaptation) implementations for efficient fine-tuning. Utilizing state-of-the-art GPUs with greater memory and compute power would enhance training efficiency, while exploring domain-specific use cases in areas like sentiment analysis or medical text classification could further demonstrate the adaptability and scalability of this hybrid framework.

## 5.3 Author's Note

The author realizes that this research is not perfect and is made with full of shortcomings. The author hopes to receive all forms of feedback and criticism from readers in order to improve the author's skills. The author also acknowledges that the use of translation and artificial intelligence software, such as DeepL, QuillBot, and ChatGPT has helped the author in writing this paper. Nevertheless, the author still pays attention to originality, ethical use of AI and retrieval of information based on credible sources so as to produce authentic work and contribute to the field of quantum machine learning. The authors are grateful to all those who have supported the research and writing of this scientific work.

## Code

github.com/kianaaa19/leopold

## Contact

santawishal17@gmail.com